\documentclass[10pt,twocolumn,letterpaper]{article}
\usepackage[OT1]{fontenc}
\usepackage[pagenumbers]{cvpr}

\usepackage{graphicx}
\usepackage{amsmath}
\usepackage{amssymb}
\usepackage{booktabs}
\usepackage{multirow}
\usepackage{colortbl}

\definecolor{ForestGreen}{rgb}{0, 0.69, 0.31}

\newcommand{\fireicon}{\raisebox{-0.15em}{\includegraphics[height=1.0em]{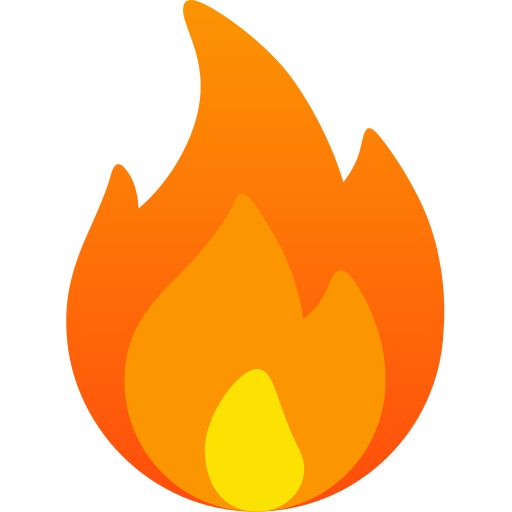}}}
\newcommand{\snowflakeicon}{\raisebox{-0.15em}{\includegraphics[height=1.0em]{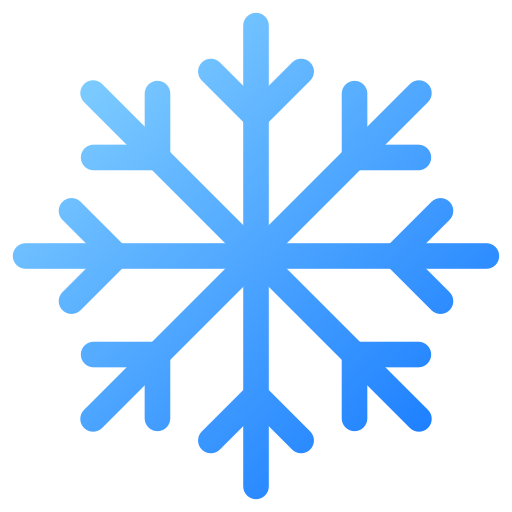}}}

\usepackage{placeins}
\usepackage{stfloats}
\usepackage{needspace}

\usepackage{float}

\makeatletter
\patchcmd{\@maketitle}{\vspace*{12pt}}{\vspace*{6pt}}{}{\PackageError{authoredcopy}{Author spacing patch failed}{}}
\patchcmd{\@maketitle}{\vspace*{24pt}}{\vspace*{12pt}}{}{\PackageError{authoredcopy}{Title spacing patch failed}{}}
\makeatother

\fnbelowfloat

\definecolor{cvprblue}{rgb}{0.21,0.49,0.74}
\usepackage[breaklinks,colorlinks,allcolors=cvprblue]{hyperref}

\def\confName{CVPR}
\def\confYear{2027}

\title{SLGNet: Unifying Structural and Language Guidance for Multimodal Object Detection}

\author{
Xiantai Xiang$^{1}$\thanks{Equal contribution.}\quad
Jiacheng Deng$^{2}$\footnotemark[1]\quad
Kai Liu$^{1}$\quad
Wenshuai Li$^{1}$\\
Zixiao Wen$^{1}$\quad
Ben Niu$^{3}$\thanks{Corresponding author.}\quad
Guangyao Zhou$^{3}$\\
$^{1}$University of the Chinese Academy of Sciences\\
$^{2}$University of Science and Technology of China\\
$^{3}$Aerospace Information Research Institute, Chinese Academy of Sciences
}

\hypersetup{
  pdftitle={SLGNet: Unifying Structural and Language Guidance for Multimodal Object Detection},
  pdfauthor={Xiantai Xiang, Jiacheng Deng, Kai Liu, Wenshuai Li, Zixiao Wen, Ben Niu, Guangyao Zhou}
}
\AtBeginDvi{}
\begin{document}
\maketitle
\raggedbottom

\begin{abstract}
RGB--infrared object detection combines complementary appearance and thermal cues to improve detection under challenging imaging conditions.
However, adapting a pretrained backbone does not by itself ensure that useful geometry is preserved during fusion or that the resulting features are effectively used across scenes.
Without explicit geometric guidance, weakened boundaries can make object regions harder to separate and compromise localization.
Even when useful geometry is retained, changes in scene context may leave target responses weak or spatially incomplete, limiting the use of available evidence.
We propose SLGNet, a parameter-efficient adaptation architecture that addresses both feature construction and use through structure-guided fusion and scene-conditioned recalibration.
The Structure-Aware Adapter uses a complementary edge reference to weight multi-scale RGB--infrared fusion and injects the resulting features into a frozen vision transformer.
Language-Guided Modulation converts structured scene descriptions from a pretrained vision-language model into channel-wise affine parameters that rescale and shift the final structure-enriched backbone features according to scene context.
We evaluate SLGNet on LLVIP, FLIR, KAIST, and DroneVehicle.
With 12.1M trainable parameters, SLGNet achieves 66.1 mAP on LLVIP and 45.1 mAP on FLIR, the highest mAP scores among the compared methods on both benchmarks.
These results improve over the reported UniRGB-IR scores by 2.9 and 1.0 points, respectively.
On FLIR, SLGNet also outperforms our full fine-tuning baseline by 1.5 mAP while using approximately 87\% fewer trainable parameters.
\end{abstract}
\section{INTRODUCTION}

RGB--infrared object detection combines complementary appearance and thermal cues to improve detection under challenging imaging conditions, with applications in autonomous driving and aerial surveillance~\cite{choi2018kaist,sun2022drone,xiang2025infrared}.
RGB images provide texture and color information, while infrared images can make targets more distinguishable when visible cues are weak~\cite{li2019illumination,liu2016multispectral}.
However, changes in illumination, thermal contrast, and scene composition alter the visibility and reliability of these cues~\cite{zhang2019cross,zhu2025wavemamba}.
Robust RGB--infrared detection therefore requires not only preserving localization-relevant geometric cues during multimodal fusion, but also adapting how the resulting target evidence is represented under varying scene conditions.

\begin{figure}[!t]
    \centering
    \includegraphics[width=\linewidth]{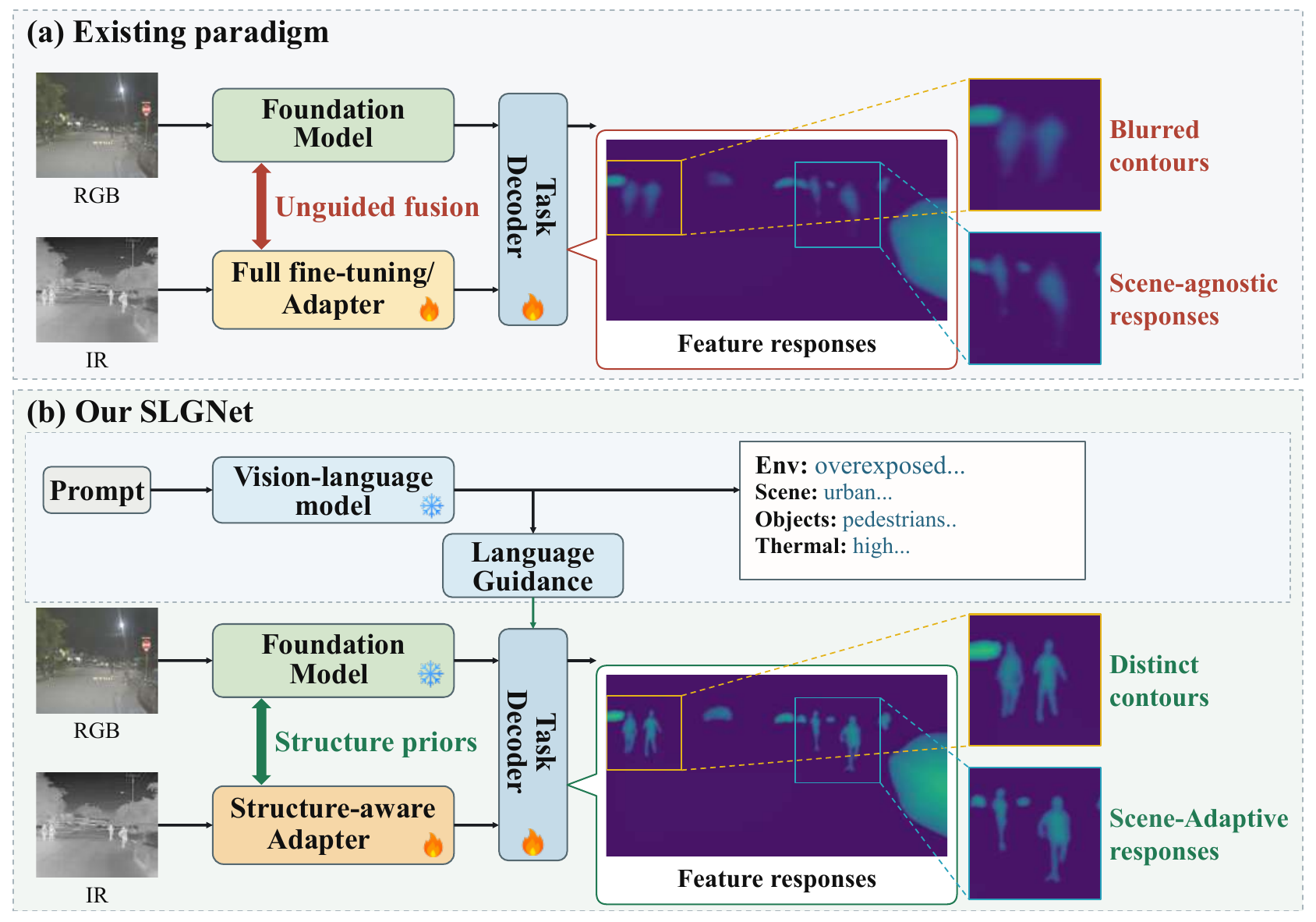}
    \caption{Comparison of RGB--infrared adaptation paradigms.
\textbf{(a)} Existing paradigms perform full fine-tuning or adapter tuning with unguided fusion.
\textbf{(b)} \textbf{SLGNet} adapts a frozen foundation model by combining a Structure-Aware Adapter for structural prior injection with VLM-driven Language-Guided Modulation for semantic feature recalibration. The \snowflakeicon\ and \fireicon\ icons indicate frozen and trainable parameters, respectively.}
    \label{fig:teaser}
    \vspace{-5mm}
\end{figure}

RGB-pretrained vision foundation models, including DINOv2~\cite{oquab2023dinov2} and SAM~\cite{kirillov2023segment}, offer transferable representations learned from large-scale visual data.
Existing approaches adapt these backbones to RGB--infrared detection through full fine-tuning or multimodal adapter tuning (Fig.~\ref{fig:teaser}(a)).
Full fine-tuning updates the backbone but incurs substantial training and task-specific storage costs.
Multimodal adapters instead inject complementary features into a frozen backbone for parameter-efficient adaptation~\cite{yuan2025unirgb}, but effective integration of thermal cues into RGB-pretrained representations remains challenging.
Regardless of the tuning strategy, fusion without explicit geometric guidance may weaken boundary cues.
As illustrated in Fig.~\ref{fig:teaser}(a), blurred boundaries can make target regions harder to separate from the background or adjacent objects, leaving object extent ambiguous and compromising localization.
Even when useful geometric cues are preserved, informative target features may remain insufficiently emphasized under varying scene conditions, such as nighttime or low-contrast environments.
As illustrated in Fig.~\ref{fig:teaser}(a), weak or distorted target responses suggest that the retained evidence is not fully exploited for detection.
Consequently, backbone adaptation alone does not ensure that fused representations retain precise object boundaries or provide reliable target evidence.

To address these challenges, we propose \textbf{SLGNet}, a parameter-efficient RGB--infrared detection framework that unifies structure-guided feature construction with scene-conditioned recalibration (Fig.~\ref{fig:teaser}(b)).
Boundary cues may be clearer in different modalities, so a shared reference that combines their complementary structure provides a common basis for weighting their contributions during fusion.
The \textbf{Structure-Aware Adapter} constructs this reference from RGB and infrared edge features and uses each modality's structural similarity to it to weight multi-scale fusion.
The fused features are injected into a frozen vision transformer through sparse attention, providing boundary-aware evidence for localization (Sec.~\ref{sec:adapter}).
While these boundary cues support localization, informative target features may still be insufficiently emphasized under varying scene conditions.
Explicit scene descriptions make these conditions available to guide channel-wise recalibration.
\textbf{Language-Guided Modulation (LGM)} uses a pretrained vision-language model~\cite{bai2025qwen2} to generate structured descriptions of illumination, scene type, object distribution, and thermal contrast.
A frozen CLIP text encoder~\cite{radford2021learning} and trainable projection heads convert these descriptions into channel-wise affine parameters~\cite{perez2018film} (Sec.~\ref{sec:guidance}).
These parameters rescale and shift the final backbone features according to scene context to improve the use of retained target evidence for detection.
Together, the two modules address complementary aspects of multimodal adaptation: structural guidance enriches the representation with explicit geometric cues relevant to object localization, while language-derived scene context guides the recalibration of this structure-enriched representation.
This coupling connects feature construction with scene-conditioned feature use while keeping the pretrained backbone frozen.

Our contributions are summarized as follows:
\begin{itemize}
    \item We propose SLGNet, a parameter-efficient RGB--infrared detection framework that integrates structural priors and language-guided modulation for geometry-aware, scene-conditioned adaptation of a frozen backbone (Sec.~\ref{method}).
    \item We design a Structure-Aware Adapter that weights multi-scale RGB--infrared fusion using a complementary edge reference and injects the fused features into a frozen vision transformer through sparse attention (Sec.~\ref{sec:adapter}).
    \item We introduce Language-Guided Modulation, which maps VLM-generated structured scene descriptions to channel-wise affine parameters that recalibrate structure-enriched features according to scene context (Sec.~\ref{sec:guidance}).
    \item Experiments on four RGB--infrared benchmarks demonstrate leading overall detection performance among the compared methods. With 12.1M trainable parameters, SLGNet uses approximately 87\% fewer trainable weights than our full fine-tuning baseline while improving accuracy on FLIR and DroneVehicle. Component ablations support the complementary roles of structural guidance and scene conditioning (Sec.~\ref{sec:experiments}).
\end{itemize}
\section{RELATED WORK}

\subsection{RGB--Infrared Object Detection}

RGB--infrared detection combines appearance and thermal cues through different fusion designs.
Liu et al.~\cite{liu2016multispectral} compare fusion at several network stages, while illumination-aware weighting~\cite{li2019illumination} and cross-modal alignment~\cite{zhang2019weakly} address changing lighting conditions and spatial discrepancies.
DAMSDet~\cite{guo2024damsdet} uses adaptive feature fusion to combine complementary information across modalities.
WaveMamba~\cite{zhu2025wavemamba} uses wavelet decomposition, with Mamba-based fusion for low-frequency features and absolute-maximum fusion for high-frequency features.

Pretrained vision transformers can also be adapted through auxiliary feature branches.
ViT-Adapter~\cite{chen2022vision} introduces convolutional spatial priors and multi-scale feature interaction for dense prediction.
UniRGB-IR~\cite{yuan2025unirgb} extracts multi-scale multimodal features and injects them into a frozen ViT through sparse attention.
SLGNet follows this frozen-backbone adaptation setting, but uses a complementary edge reference to guide fusion before feature injection.
Language-derived scene conditions then recalibrate the final structure-enriched backbone features.

\subsection{Language-Guided Detection and Feature Modulation}

Language provides different forms of guidance for visual recognition and detection.
CLIP~\cite{radford2021learning} learns image--text representations that support recognition from textual descriptions.
GLIP~\cite{li2022grounded} connects object detection with phrase grounding, using text to specify the objects to locate.
For multispectral pedestrian detection, MSCoTDet~\cite{kim2025mscotdet} generates object descriptions and combines language-driven predictions with vision-based detections.
These methods use language to describe targets or combine detection outputs.

Language can also condition visual features directly.
FiLM~\cite{perez2018film} defines feature-wise affine transformations whose scaling and bias depend on conditioning information.
LGM uses this form of modulation with structured scene descriptions generated by Qwen2.5-VL~\cite{bai2025qwen2}.
The descriptions provide conditions such as illumination and thermal contrast for recalibrating the final backbone features.
LGM therefore uses structured scene descriptions to recalibrate structure-enriched features before detection, rather than specify target classes or combine predicted boxes.

\section{THE PROPOSED METHOD}
\label{method}
\begin{figure*}[t]
    \centering
    \includegraphics[width=\textwidth]{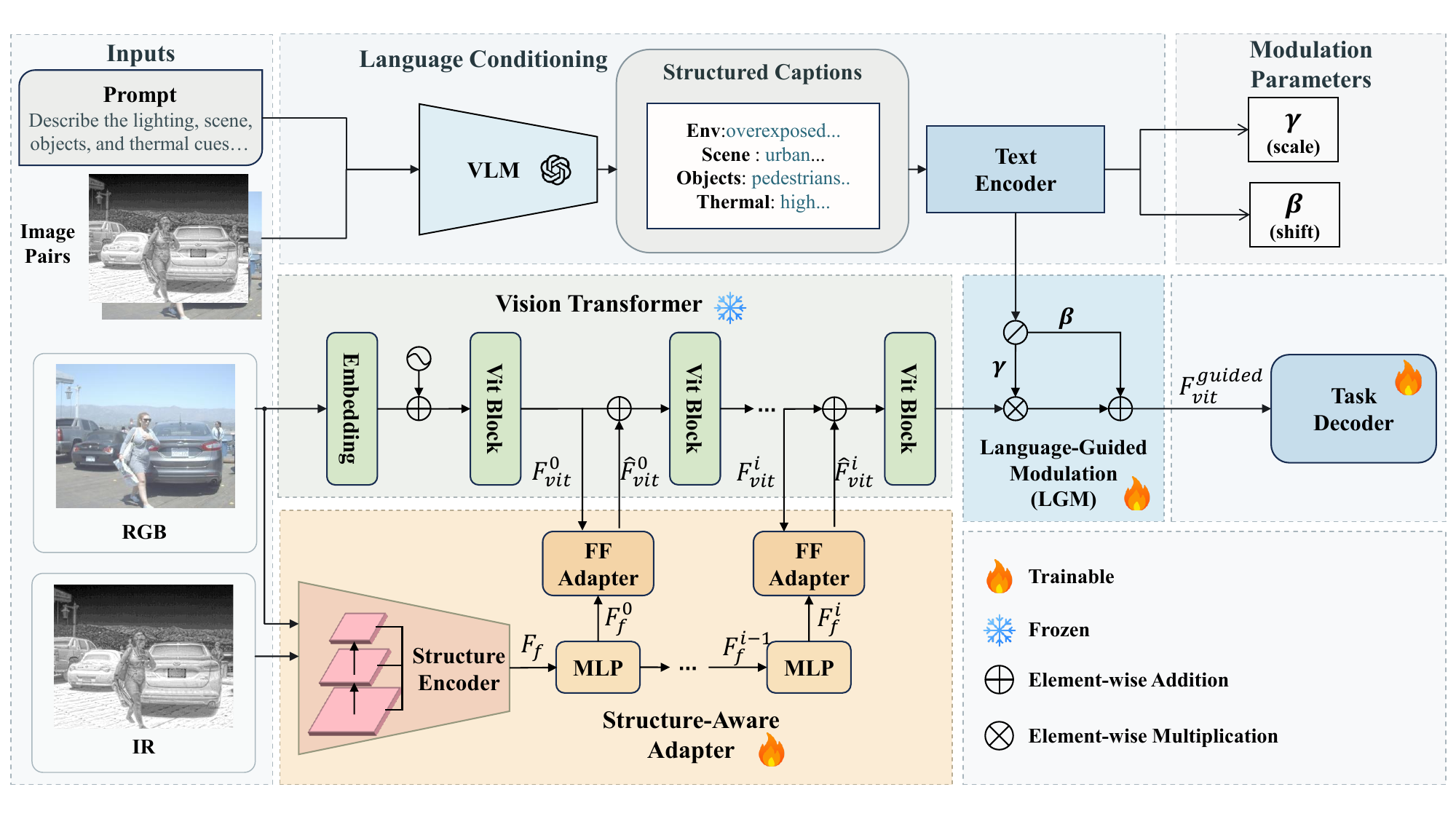}
    \caption{Overview of \textbf{SLGNet}, which pairs a frozen ViT backbone with two lightweight trainable modules: (1) the \textbf{Structure-Aware Adapter} (bottom), extracting hierarchical structural priors from paired images via a Structure Encoder and injecting them into ViT blocks via the Feature Fusion Adapter (FF-Adapter); and (2) \textbf{Language-Guided Modulation (LGM)} (right), utilizing VLM-generated structured captions (Environment, Scene, Objects, Thermal) to recalibrate the final feature map via affine transformations ($\gamma, \beta$). The \snowflakeicon\ and \fireicon\ icons denote frozen and trainable parameters, respectively.}
    \label{fig:overview}
\end{figure*}

As shown in Fig.~\ref{fig:overview}, SLGNet adapts a frozen vision transformer (ViT) for RGB--infrared object detection through structure-guided fusion and scene-conditioned feature recalibration.
The Structure-Aware Adapter uses a complementary edge reference to guide multi-scale fusion and injects the fused features into the backbone.
Language-Guided Modulation (LGM) then uses structured scene descriptions to rescale and shift the final structure-enriched backbone features before detection.

Given an RGB--infrared image pair, the RGB image is converted into patch tokens and processed by the frozen ViT.
The Structure-Aware Adapter (Sec.~\ref{sec:adapter}) extracts and fuses multi-scale features from both images, then injects them into the ViT blocks through the Feature Fusion Adapter (FF-Adapter).
A pretrained VLM generates structured scene descriptions from the same image pair.
LGM (Sec.~\ref{sec:guidance}) encodes these descriptions and predicts channel-wise scaling and bias parameters to modulate the final structure-enriched features.
The detection head uses the modulated features to predict object classes and bounding boxes.
We train the adaptation modules using the detection loss and keep the ViT and text encoder frozen.

\subsection{Structure-Aware Adapter}
\label{sec:adapter}

The Structure-Aware Adapter (SA-Adapter) uses a complementary edge reference to guide RGB--infrared feature fusion.
Its Structure Encoder (S-Encoder) extracts multi-scale features and weights the two modalities by their structural consistency with this reference.
The FF-Adapter injects the fused features into the frozen ViT.

\subsubsection{Structure Encoder}

\begin{figure}[t]
    \centering
    \includegraphics[width=\linewidth]{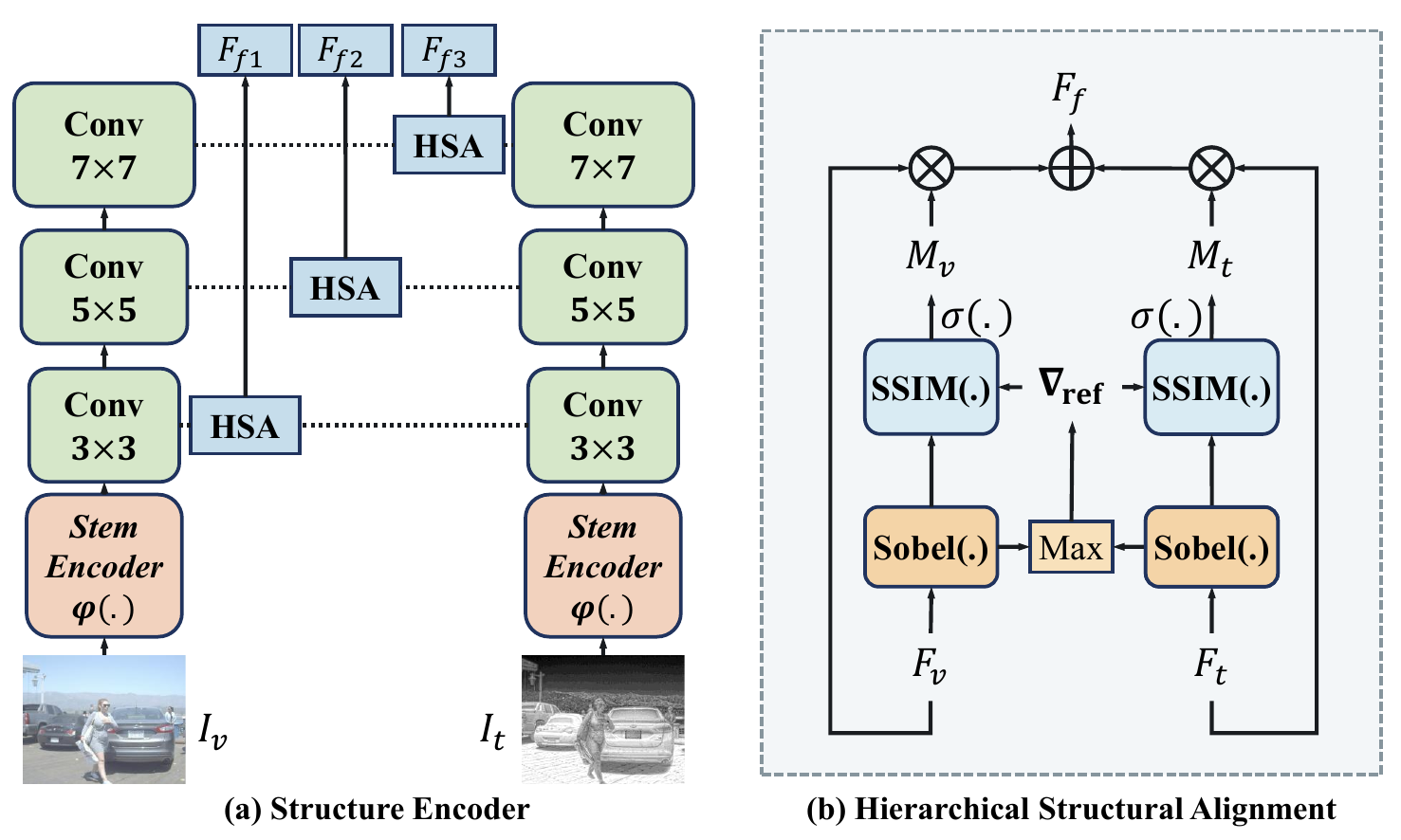}
    \caption{Architecture of the Structure Encoder. \textbf{(a)} Progressive convolutional stages extract multi-resolution structural priors. \textbf{(b)} The Hierarchical Structural Alignment (HSA) module establishes a reference map $\nabla_{\text{ref}}$ and uses an SSIM-driven mechanism to dynamically weight multimodal features.}
    \label{fig:s_encoder}
    \vspace{-5mm}
\end{figure}

As shown in Fig.~\ref{fig:s_encoder}(a), a shared stem encoder $\varphi(\cdot)$ extracts features from the RGB image $I_v$ and infrared image $I_t$.
Three convolutional stages with $3 \times 3$, $5 \times 5$, and $7 \times 7$ kernels produce feature maps $F_{vl}$ and $F_{tl}$ at $1/8$, $1/16$, and $1/32$ resolutions, respectively, where $l \in \{1,2,3\}$.
At each scale, both modality features have shape $C_l \times H_l \times W_l$, where $C_l$ is the number of channels and $H_l$ and $W_l$ denote the spatial height and width, respectively.

The Hierarchical Structural Alignment (HSA) module builds a common edge reference from both modalities (Fig.~\ref{fig:s_encoder}(b)).
At each scale $l$, it computes Sobel gradient magnitudes for each modality $m \in \{v,t\}$:
\begin{equation}
\nabla F_{ml} = \operatorname{Sobel}(F_{ml}).
\end{equation}
The element-wise maximum of the two gradient maps forms the common edge reference map:
\begin{equation}
\nabla_{\mathrm{ref},l} = \max(\nabla F_{vl}, \nabla F_{tl}).
\end{equation}
This reference retains the stronger edge response at each location, allowing edges from either modality to contribute.

HSA uses structural consistency with the reference to determine the fusion weights.
HSA computes the scores separately at each scale; the scale index on the statistics is omitted below for readability.
The SSIM-style score for modality $m$ against the edge reference is
\begin{equation}
M'_{ml} = \frac{(2\mu_m\mu_{\mathrm{ref}} + \xi_1)(2\sigma_{m,\mathrm{ref}} + \xi_2)}{(\mu_m^2 + \mu_{\mathrm{ref}}^2 + \xi_1)(\sigma_m^2 + \sigma_{\mathrm{ref}}^2 + \xi_2)},
\end{equation}
where $\mu$ and $\sigma^2$ denote means and variances, $\sigma_{m,\mathrm{ref}}$ is the covariance between the modality and reference representations, and $\xi_1$ and $\xi_2$ are numerical stability constants.
A sigmoid maps each score to a weight, $M_{ml} = \operatorname{sigmoid}(M'_{ml})$, and the weighted fused feature is
\begin{equation}
F_{f_l} = M_{vl} \odot F_{vl} + M_{tl} \odot F_{tl}.
\end{equation}
Here, $\odot$ denotes element-wise multiplication with broadcasting where needed.
The two sigmoid gates are computed independently and are not constrained to sum to one.
The edge reference guides the weighting, while the weighted sum combines the modality features rather than the gradient maps.
A $1 \times 1$ convolution projects each fused feature to the ViT embedding dimension $D$, producing $D \times H_l \times W_l$ maps for injection into the backbone.

\subsubsection{Feature Fusion Adapter}

Before each ViT block, the FF-Adapter injects the structure-guided fusion features using multi-scale deformable attention (MSDA)~\cite{zhu2020deformable}.
At block $i$, $F^{(i)}_{\mathrm{vit}} \in \mathbb{R}^{N\times D}$ contains $N$ spatial patch tokens that provide the queries; the three projected fusion maps provide the keys and values.
The attention output is added to the tokens through a residual update:
\begin{equation}
\hat{F}^{(i)}_{\mathrm{vit}} = F^{(i)}_{\mathrm{vit}} + \operatorname{MSDA}\left(F^{(i)}_{\mathrm{vit}}, \{F^{(i)}_{f_l}\}_{l=1}^{3}\right).
\end{equation}
For each query, its spatial coordinate is mapped to each feature level, and learned offsets and attention weights determine the sampled locations and their contributions.
MSDA returns an $N\times D$ update, so the residual injection preserves the token sequence shape.
The updated tokens enter the next frozen ViT block, while an MLP updates the multi-scale features between consecutive injection steps.

\subsection{Language-Guided Modulation}
\label{sec:guidance}

LGM uses scene descriptions to recalibrate the final structure-enriched backbone features.
It adjusts channel responses through a channel-wise affine transformation conditioned on the scene descriptions.

Given the RGB--infrared image pair $(I_v,I_t)$, Qwen2.5-VL~\cite{bai2025qwen2} generates four description fields $\{s_i\}_{i\in\mathcal{S}}$, where $\mathcal{S}=\{\mathrm{env},\mathrm{type},\mathrm{obj},\mathrm{therm}\}$.
The fields describe illumination and weather, scene type, target presence and distribution, and relative infrared contrast, respectively.
This format separates different aspects of the current scene rather than limiting the text to target class names.

A frozen CLIP text encoder~\cite{radford2021learning} encodes each field as $E_i = \mathcal{CLIP}_{\text{text}}(s_i) \in \mathbb{R}^{L \times d}$, where $L$ is the token sequence length and $d$ is the embedding dimension. The four embeddings are concatenated in the order \texttt{env}, \texttt{type}, \texttt{obj}, and \texttt{therm} along the channel dimension, giving an $L \times 4d$ tensor. A trainable MLP projects each token to $d$ dimensions, and pooling aggregates the resulting $L \times d$ tensor into a compact $d$-dimensional scene vector $h \in \mathbb{R}^d$:
\begin{equation}
h=\operatorname{Pool}\left(\operatorname{MLP}_{\mathrm{proj}}\left(\operatorname{Concat}_{i\in\mathcal{S}} E_i\right)\right).
\end{equation}
Here, $\operatorname{Pool}$ aggregates the token dimension, giving $h\in\mathbb{R}^{d}$.
Two trainable projection heads, $\operatorname{MLP}_{\gamma}$ and $\operatorname{MLP}_{\beta}$, map $h$ to scaling and bias vectors $\gamma,\beta\in\mathbb{R}^{C}$.

The final spatial tokens are reshaped into $F_{\mathrm{vit}}\in\mathbb{R}^{C\times H\times W}$, where $C=D$ and $HW=N$.
Following feature-wise affine conditioning~\cite{perez2018film}, LGM applies
\begin{equation}
F_{\mathrm{vit}}^{\mathrm{guided}}=(1+\gamma)\odot F_{\mathrm{vit}}+\beta.
\end{equation}
The vectors $\gamma$ and $\beta$ are reshaped to $C\times 1\times 1$ and broadcast over all $H\times W$ spatial positions.
Each image pair therefore uses one shared set of channel parameters across its feature map.
The transform reduces to the identity when $\gamma=0$ and $\beta=0$.
LGM predicts channel parameters without generating spatial attention maps. The modulated features are fed into the detection head.

\section{EXPERIMENTS}
\label{sec:experiments}
\raggedbottom
\setlength{\intextsep}{6pt plus 2pt minus 2pt}
\setlength{\textfloatsep}{12pt plus 2pt minus 2pt}

\begin{table*}[!t]
    \centering
    \small
    \setlength{\tabcolsep}{5pt}
    \caption{Detection results on LLVIP and FLIR. Bold and underlined values mark the best and second-best results, including ties. Higher AP is better; fewer trainable parameters are better. Trainable parameters count only weights updated during training, not the total model size or inference cost. \textquotedblleft\textendash\textquotedblright{} indicates an unreported or otherwise unavailable value.}
    \label{tab:LLVIP}
  
    \newcommand{\missingvalue}[1]{\begingroup\setbox0=\hbox{#1}\makebox[\wd0][c]{\textendash}\endgroup}
    \begin{tabular}{l c rrr rrr r}
    \toprule
    \multirow{2}{*}{\textbf{Methods}} & \multirow{2}{*}{\textbf{Modality}} & 
    \multicolumn{3}{c}{\textbf{LLVIP}} &
    \multicolumn{3}{c}{\textbf{FLIR}} &
    \multirow{2}{*}{\textbf{Trainable Params}} \\
    \cmidrule(lr){3-5} \cmidrule(lr){6-8}
    &  & \textbf{mAP} & \textbf{mAP}\textsubscript{50} & \textbf{mAP}\textsubscript{75} 
    & \textbf{mAP} & \textbf{mAP}\textsubscript{50} & \textbf{mAP}\textsubscript{75} & \\
    \midrule
  
    FasterRCNN~\cite{ren2015faster}       & IR       & 54.5 & 94.6 & 57.6 & 37.6 & 75.8 & 31.6 & 68.5M \\
    RetinaNet~\cite{lin2017focal}        & IR       & 55.1 & 94.8 & 57.6 & 31.5 & 66.1 & 25.3 & 43.0M\\
    YOLOV8~\cite{Jocher_Ultralytics_YOLO_2023}           & IR       & 62.1 & 95.2 & 67.0 & 38.3 & 72.9 & 31.8 & 76.7M \\
    DDQ-DETR~\cite{zhang2023dense}         & IR       & 58.6 & 93.9 & 64.6 & 37.1 & 73.9 & 32.2 & 244.6M \\
    \midrule
  
    FasterRCNN~\cite{ren2015faster}       & RGB      & 45.1 & 87.0 & 41.2 & 27.7 & 62.2 & 21.2 & 68.5M\\
    RetinaNet~\cite{lin2017focal}        & RGB      & 42.8 & 88.0 & 34.4 & 21.9 & 51.2 & 15.2 & 43.0M \\
    YOLOV8~\cite{Jocher_Ultralytics_YOLO_2023}           & RGB      & 54.0 & 91.9 & 52.5 & 28.2 & 66.3 & 24.2 & 76.7M \\
    DDQ-DETR~\cite{zhang2023dense}         & RGB      & 46.7 & 86.1 & 45.8 & 30.9 & 64.9 & 24.5 & 244.6M \\
  
    \midrule
  
    ICAFusion~\cite{shen2024icafusion}     & RGB+IR   & \missingvalue{00.0} & \missingvalue{00.0} & \missingvalue{00.0} & 41.4 & 79.2 & 36.9 & 120.0M \\
    RSDet~\cite{zhao2024removal}         & RGB+IR   & 61.3 & 95.8 & 70.4 & 43.8 & \underline{83.9} & 40.1 & 386.3M\\
    UniRGB-IR~\cite{yuan2025unirgb}     & RGB+IR   & 63.2 & 96.1 & 72.2 & 44.1 & 81.4 & 40.2 & \textbf{8.9M} \\
    CrossModalNet~\cite{li2025crossmodalnet} & RGB+IR   & 64.7 & 97.7 & 73.5 & 43.3 & 81.7 & 39.1 & 92.8M \\
    COFNet~\cite{zhou2025cofnet}    & RGB+IR   & 65.9 & 97.7 & \underline{75.9} & \underline{44.6} & 83.6 & \underline{41.7} & 90.2M \\
    
    SFEDet~\cite{tian2026sfedet} & RGB+IR & 65.9 & 96.8 & \textbf{77.1} & 43.0 & 81.7 & 38.9 & 24.5M \\
    FreDFT~\cite{wu2026fredft} & RGB+IR & \underline{66.0} & \underline{98.2} & 74.7 & 42.6 & 83.5 & 37.3 & 152.6M \\
    \rowcolor[gray]{0.94} SLGNet (Ours) & RGB+IR   & \textbf{66.1} & \textbf{98.3} & 75.4 & \textbf{45.1} & \textbf{85.8} & \textbf{42.3} & \underline{12.1M} \\
    \bottomrule
    \end{tabular}
  
\end{table*}

\subsection{Experimental Setup}
\noindent\textbf{Datasets.} We evaluate SLGNet on four RGB--infrared benchmarks.
LLVIP~\cite{jia2021llvip} provides low-light pedestrian image pairs, with 12,025 for training and 3,463 for testing.
FLIR~\cite{zhang2020multispectral} contains 4,129 training and 1,013 test pairs of aligned RGB--infrared driving images across three classes: person, car, and bicycle.
KAIST~\cite{hwang2015multispectral} contains 95,328 visible--thermal pairs for pedestrian detection.
We report overall (All), daytime (Day), and nighttime (Night) results on its 2,252-pair test split.
DroneVehicle~\cite{sun2022drone} provides paired aerial RGB--infrared images with oriented bounding-box annotations for five vehicle classes.

\noindent\textbf{Evaluation Metrics.} For LLVIP, FLIR, and DroneVehicle, we report mean average precision (mAP) and $\text{mAP}_{50}$, with $\text{mAP}_{75}$ additionally reported for LLVIP and FLIR. Rotated IoU is used for DroneVehicle. For KAIST, we report the log-average miss rate ($\text{MR}^{-2}$) over the false positives per image (FPPI) range of $[10^{-2}, 10^0]$. Higher mAP scores and lower $\text{MR}^{-2}$ indicate better performance.

\noindent\textbf{Implementation Details.} We use MMDetection for horizontal bounding box detection on LLVIP, FLIR, and KAIST, and MMRotate for oriented object detection on DroneVehicle. We adopt a frozen ViT-Base backbone initialized with DINOv2~\cite{oquab2023dinov2} pre-trained weights and insert the Structure-Aware Adapter before all 12 transformer blocks. Models are trained on NVIDIA H20 GPUs for 50 epochs with a batch size of 8 and Automatic Mixed Precision (AMP), using the AdamW optimizer with an initial learning rate of $1 \times 10^{-4}$, a weight decay of 0.1, and a layer-wise learning rate decay factor of 0.7.

\subsection{Comparisons with State-of-the-Art Methods}

We compare SLGNet with representative RGB--infrared detectors on four benchmarks.
Table~\ref{tab:LLVIP} also reports RGB-only and infrared-only results for reference.

\Needspace*{2\baselineskip}
\noindent\textbf{LLVIP and FLIR.} Table~\ref{tab:LLVIP} shows that SLGNet obtains the highest mAP among the compared methods, with 66.1 on LLVIP and 45.1 on FLIR.
It also has the highest LLVIP $\mathrm{mAP}_{50}$ and the highest scores on all three FLIR metrics, while LLVIP $\mathrm{mAP}_{75}$ is 75.4 compared with COFNet's 75.9.
Relative to UniRGB-IR, mAP increases by 2.9 and 1.0 points with 3.2M additional trainable parameters.
Compared with COFNet, SLGNet improves mAP by 0.2 and 0.5 points while using 86.6\% fewer trainable parameters.

\begin{table}[!htbp]
    \centering
    \small
    \caption{KAIST pedestrian detection results. Lower $\mathrm{MR}^{-2}$ is better. Bold and underlined values mark the best and second-best results, respectively. \textit{Handcrafted} denotes methods that use hand-designed features without a neural backbone.}
    \label{tab:KAIST}
    \setlength{\tabcolsep}{3pt}
    \begin{tabular}{l c rrr}
    \toprule
    \multirow{2}{*}{\textbf{Methods}} & \multirow{2}{*}{\textbf{Backbone}} &  \multicolumn{3}{c}{$\mathrm{MR}^{-2}$ (\%) $\downarrow$} \\
    \cmidrule(lr){3-5} 
     &  & \textbf{All} & \textbf{Day} & \textbf{Night} \\
    \midrule
    ACF~\cite{hwang2015multispectral}           & Handcrafted    & 67.74 & 64.31 & 75.06 \\
    HalfwayFusion~\cite{liu2016multispectral} & VGG-16    & 49.18 & 47.58 & 52.35 \\
    IATDNN+IASS~\cite{guan2019fusion}   & VGG-16    & 48.96 & 49.02 & 49.37 \\
    CIAN~\cite{zhang2019cross}          & VGG-16    & 35.53 & 36.02 & 32.38 \\
    AR-CNN~\cite{zhang2019weakly}        & VGG-16    & 34.95 & 34.36 & 36.12 \\
    \midrule
    MBNet~\cite{zhou2020improving}         & ResNet-50 & 31.87 & 32.37 & 30.95 \\
    CMPD~\cite{li2022confidence}          & ResNet-50 & 28.98 & 28.30 & 30.56 \\
    CAGTDet~\cite{yuan2024improving}       & ResNet-50 & 28.96 & 28.79 & 27.73 \\
    C$^2$Former~\cite{yuan2024c2former}      & ResNet-50 & 28.39 & 28.48 & 26.67 \\
    \midrule
    UniRGB-IR~\cite{yuan2025unirgb}        & ViT-B     & 25.21 & 25.93 & 23.95 \\
    M-SpecGene~\cite{zhou2025m}    & ViT-B     & \underline{23.74} & \underline{25.66} & \textbf{19.42} \\
    \rowcolor[gray]{0.94} SLGNet (Ours)    & ViT-B     & \textbf{19.88} & \textbf{21.01} & \underline{20.56} \\
    \bottomrule
    \end{tabular}
\end{table}

\Needspace*{2\baselineskip}
\noindent\textbf{KAIST.} SLGNet has the lowest overall $\mathrm{MR}^{-2}$ of 19.88\% in Table~\ref{tab:KAIST}, 3.86 and 5.33 percentage points below M-SpecGene and UniRGB-IR, respectively.
The daytime miss rate is also the lowest at 21.01\%, whereas the nighttime rate is 20.56\%, compared with 19.42\% for M-SpecGene.

\begin{table}[!htbp]
    \caption{DroneVehicle oriented detection results. Higher $\mathrm{mAP}_{50}$ and per-class $\mathrm{AP}_{50}$ are better. Bold and underlined values mark the best and second-best results, respectively. }
    \centering
    \small
    \setlength{\tabcolsep}{2pt}
    \begin{tabular}{l r rrrrr}
    \toprule
    \textbf{Methods} & \textbf{mAP}\textsubscript{50} & \textbf{Car} & \textbf{Truck} & \shortstack{\textbf{Freight-}\\\textbf{Car}} & \textbf{Bus} & \textbf{Van} \\
    \midrule  
    Halfway Fusion~\cite{liu2016multispectral}     & 70.0  & 90.1  & 62.3  & 58.5  & 89.1  & 49.8   \\
    MBNet\footnotemark~\cite{zhou2020improving}               & 71.9 & 90.1  & 64.4  & 62.4  & 88.8  & 53.6  \\
    TSFADet~\cite{yuan2022translation}             & 73.1 & 89.9  & 67.9  & 63.7  & 89.8  & 54.0   \\
    C$^2$Former~\cite{yuan2024c2former}            & 74.2 & 90.2  & 68.3  & 64.4  & 89.8  & 58.5  \\
    AFFCM~\cite{wu2023vehicle}              & 76.6  & 90.2  & 73.4  & 64.9  & 89.9  & 64.9 \\
    MC-DETR~\cite{ouyang2023multi}            & 76.9 & 94.8  & 76.7  & 60.4  & 91.1  & 61.4   \\
    M2FP~\cite{ouyang2024multimodal}                & 78.7 & 95.7  & 76.2  & 64.7  & \textbf{92.1} & 64.7  \\
    DMM~\cite{zhou2025dmm} & 79.4  & 90.4 & 79.8 & 68.2  & 89.9  & \textbf{68.6} \\
    UniFusOD~\cite{xiang2025infrared} & 79.5  & \textbf{96.4} & \textbf{81.3} & 63.5  & 90.8  & \underline{65.6} \\
    WaveMamba~\cite{zhu2025wavemamba} & \underline{79.8}  & 95.0 & 80.4 & \underline{68.5}  & 90.6  & 64.5 \\
    \rowcolor[gray]{0.94} SLGNet (Ours) & \textbf{80.7}  & \underline{96.1} & \underline{80.9} & \textbf{69.4}  & \underline{91.8}  & 65.3 \\
    \bottomrule
    \end{tabular}
    \label{tab:DroneVehicle_comparison}
\end{table}

\Needspace*{2\baselineskip}
\noindent\textbf{DroneVehicle.} SLGNet reaches 80.7 $\mathrm{mAP}_{50}$ in Table~\ref{tab:DroneVehicle_comparison}, the highest value among the compared methods and 0.9 points above WaveMamba.
Its Freight-Car $\mathrm{AP}_{50}$ exceeds WaveMamba's by 0.9 points, while its Van $\mathrm{AP}_{50}$ is 65.3, below UniFusOD (65.6) and DMM (68.6).

\subsection{Ablation Study}

\subsubsection{Component Contributions}

To assess the contribution of the SA-Adapter and the additional benefit of LGM, we add the modules sequentially on FLIR and DroneVehicle, as summarized in Table~\ref{tab:ablation_component}.
The baseline concatenates RGB and infrared inputs at the pixel level; only its patch embedding and detection head are trainable, while the remaining ViT parameters are frozen.

\begin{table}[!b]
    \centering
    \small
    \setlength{\tabcolsep}{3pt}
    \caption{Incremental component results on FLIR and DroneVehicle. SA denotes the SA-Adapter. $\Delta$ is the gain of the full configuration over the baseline.}
    \label{tab:ablation_component}
    \begin{tabular}{l rrrr}
    \toprule
    \multirow{2}{*}{\textbf{Configuration}} &
    \multicolumn{2}{c}{\textbf{FLIR}} &
    \multicolumn{2}{c}{\textbf{DroneVehicle}} \\
    \cmidrule(lr){2-3} \cmidrule(lr){4-5}
    & $\mathrm{mAP}$ & $\mathrm{mAP}_{50}$ & $\mathrm{mAP}$ & $\mathrm{mAP}_{50}$ \\
    \midrule
    Baseline & 42.3 & 79.7 & 53.8 & 76.7 \\
    Baseline + SA & 44.3 & 82.4 & 55.1 & 78.6 \\
    \rowcolor[gray]{0.94} Baseline + SA + LGM & \textbf{45.1} & \textbf{85.8} & \textbf{57.2} & \textbf{80.7} \\
    \midrule
    $\Delta$ & +2.8 & +6.1 & +3.4 & +4.0 \\
    \bottomrule
    \end{tabular}
\end{table}

Adding the SA-Adapter increases mAP by 2.0 points on FLIR and 1.3 points on DroneVehicle.
LGM provides a further 0.8 and 2.1 points, respectively, bringing the total gains over the baseline to 2.8 and 3.4 points.
The additional gains with LGM support the use of scene-conditioned recalibration on top of the structure-enriched representation.

\subsubsection{Qualitative Analysis of the SA-Adapter}

\begin{figure}[tbp]
    \centering
    \includegraphics[width=\columnwidth]{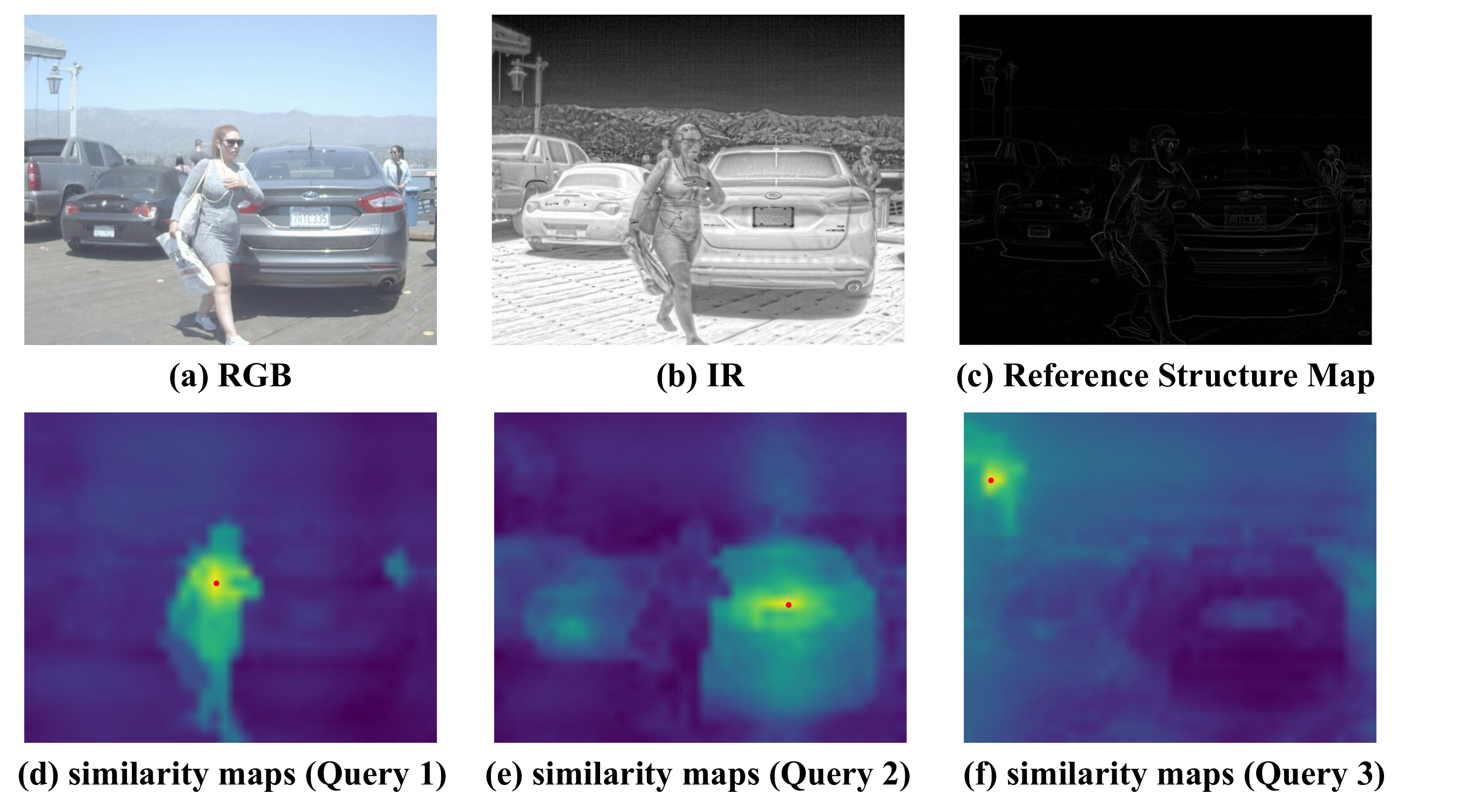}
    \caption{Query-based feature similarity of backbone representations enriched by the SA-Adapter. (a)--(b) RGB and infrared inputs. (c) Complementary edge reference $\nabla_{\mathrm{ref}}$. (d)--(f) Cosine similarity between the feature at each query patch, marked by a red dot, and all output tokens. High-similarity regions follow the pedestrian, vehicle, and street-light structures.}
    \label{fig:vis_structure}
\end{figure}

We examine the spatial organization of the backbone features enriched by the SA-Adapter through the similarity maps in Fig.~\ref{fig:vis_structure}.
For each query patch marked by a red dot, we compute its cosine similarity with all output tokens.
The RGB and infrared inputs and the complementary edge reference provide context for reading these maps.
\footnotetext{The MBNet results are from the DroneVehicle reimplementation reported in Table~IV of C$^2$Former~\cite{yuan2024c2former}.}

In Fig.~\ref{fig:vis_structure}(d)--(e), high-similarity regions follow the pedestrian and vehicle contours, with weaker responses in the surrounding background.
For the query on an unannotated street light in Fig.~\ref{fig:vis_structure}(f), the similarity map extends along the pole.
These spatial patterns are consistent with the SA-Adapter's structure-guided fusion and feature injection, which provide geometric cues to the frozen backbone.

\Needspace*{3\baselineskip}
\subsubsection{Analysis of Language-Guided Modulation}

\noindent\textbf{Scene Descriptions.} We compare three text formats to examine the role of scene information in LGM, as summarized in Table~\ref{tab:prompt_ablation}.
Class lists provide target names, and free-form captions describe the scene without fixed fields.
Structured captions describe illumination, scene type, object distribution, and thermal contrast in separate fields.

Free-form captions improve mAP over class lists by 0.8 points on FLIR and 0.9 on DroneVehicle, showing the benefit of scene information beyond target names.
Structured captions provide a further 0.4 and 0.9 points, reaching 45.1 and 57.2 mAP, respectively.
The corresponding $\mathrm{mAP}_{50}$ gains over free-form captions are 1.3 and 0.8 points.
Compared with class lists, the structured format adds 1.2 and 1.8 mAP points and 3.8 and 1.9 $\mathrm{mAP}_{50}$ points.
These gains across both datasets support using structured scene descriptions to condition feature recalibration.

\begin{table}[H]
    \centering
    \small
    \setlength{\tabcolsep}{3pt}
    \caption{LGM text-format comparison on FLIR and DroneVehicle. Structured captions follow Sec.~\ref{sec:guidance}; bold marks the highest scores.}
    \label{tab:prompt_ablation}
    \begin{tabular}{l rrrr}
    \toprule
    \multirow{2}{*}{\textbf{Text Format}} &
    \multicolumn{2}{c}{\textbf{FLIR}} &
    \multicolumn{2}{c}{\textbf{DroneVehicle}} \\
    \cmidrule(lr){2-3} \cmidrule(lr){4-5}
    & $\mathrm{mAP}$ & $\mathrm{mAP}_{50}$ & $\mathrm{mAP}$ & $\mathrm{mAP}_{50}$ \\
    \midrule
    Class List & 43.9 & 82.0 & 55.4 & 78.8 \\
    Free-form Caption & 44.7 & 84.5 & 56.3 & 79.9 \\
    \rowcolor[gray]{0.94} Structured Caption & \textbf{45.1} & \textbf{85.8} & \textbf{57.2} & \textbf{80.7} \\
    \bottomrule
    \end{tabular}
\end{table}

\Needspace*{5\baselineskip}
\raggedbottom
\noindent\textbf{Text Encoders.} We compare five pretrained text encoders for LGM on FLIR and DroneVehicle to examine the choice of representation for scene conditioning.
Table~\ref{tab:text_encoder} compares BERT, T5, and RoBERTa with the vision--language models BLIP-2 and CLIP.
This experiment complements the text-format comparison by examining how scene descriptions are encoded for channel-wise feature modulation.

\begin{table}[H]
    \centering
    \small
    \setlength{\tabcolsep}{3pt}
    \caption{Text encoder comparison for LGM on FLIR and DroneVehicle. Higher AP is better; bold marks the highest scores.}
    \label{tab:text_encoder}
    \begin{tabular}{l rrrr}
    \toprule
    \multirow{2}{*}{\textbf{Text Encoder}} &
    \multicolumn{2}{c}{\textbf{FLIR}} &
    \multicolumn{2}{c}{\textbf{DroneVehicle}} \\
    \cmidrule(lr){2-3} \cmidrule(lr){4-5}
    & $\mathrm{mAP}$ & $\mathrm{mAP}_{50}$ & $\mathrm{mAP}$ & $\mathrm{mAP}_{50}$ \\
    \midrule
    BERT~\cite{devlin2019bert} & 43.2 & 81.9 & 51.4 & 72.5 \\
    T5~\cite{ni2022sentence} & 43.8 & 83.0 & 51.8 & 73.6 \\
    RoBERTa~\cite{liu2019roberta} & 43.6 & 82.2 & 52.2 & 73.6 \\
    BLIP-2~\cite{li2023blip} & 44.9 & 84.8 & 56.1 & 79.8 \\
    \rowcolor[gray]{0.94} CLIP~\cite{radford2021learning} & \textbf{45.1} & \textbf{85.8} & \textbf{57.2} & \textbf{80.7} \\
    \bottomrule
    \end{tabular}
\end{table}

BLIP-2 and CLIP give higher scores than the three text-only encoders on both datasets.
CLIP reaches 45.1 mAP on FLIR and 57.2 on DroneVehicle, exceeding BLIP-2 by 0.2 and 1.1 points, respectively.
Its corresponding $\mathrm{mAP}_{50}$ gains over BLIP-2 are 1.0 and 0.9 points.
These results support the use of CLIP to encode scene conditions for LGM.

\begin{figure}[tbp]
    \centering
    \includegraphics[width=\columnwidth]{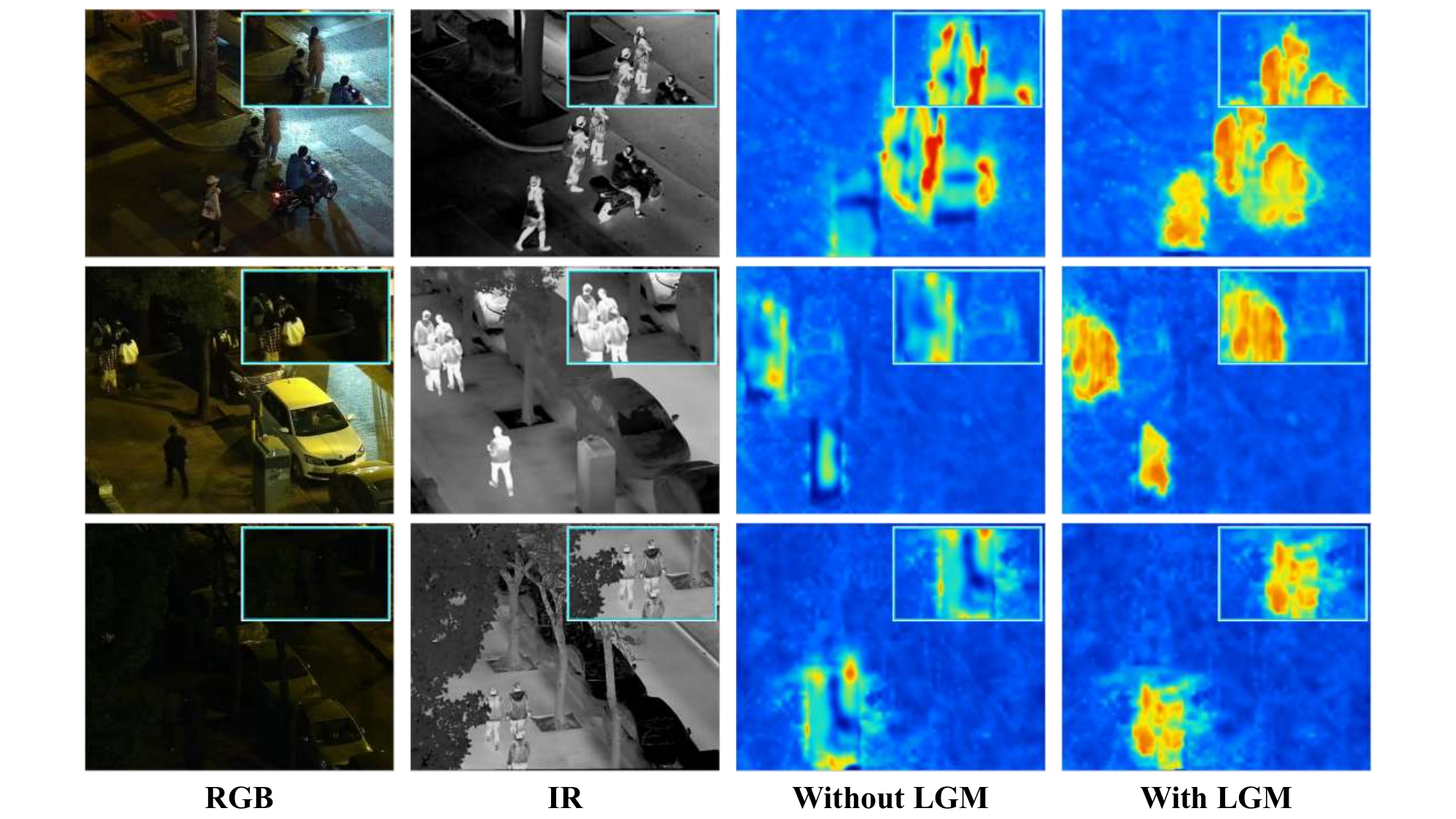}
    \caption{Feature responses with and without LGM under low illumination. Columns show RGB and infrared inputs, followed by response maps without and with LGM. Responses cover pedestrians and riders more coherently with LGM. Cyan insets show enlarged target regions.}
    \label{fig:lgm_vis}
\end{figure}

\begin{figure}[!htbp]
    \centering
    \includegraphics[width=0.88\columnwidth]{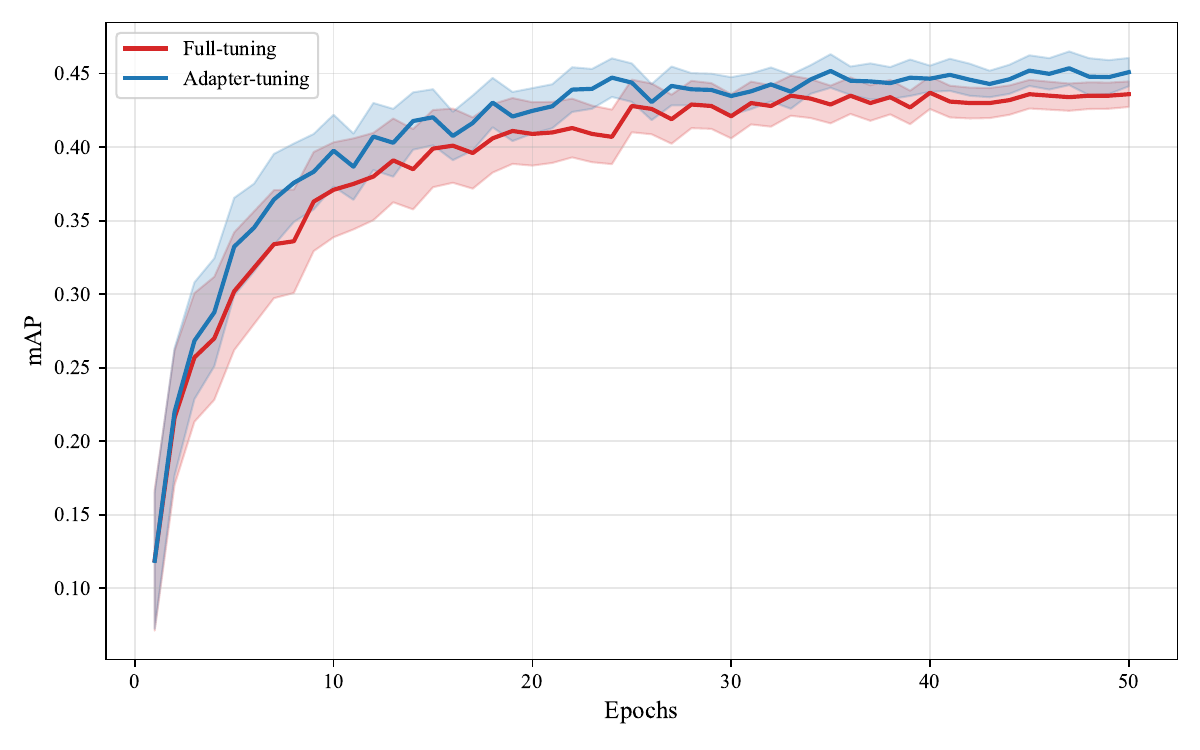}
    \caption{Validation mAP during training on FLIR. Blue and red denote adapter tuning and full fine-tuning, respectively. Solid lines show the mean, and shaded bands show the standard deviation.}
    \label{fig:training_curve}
\end{figure}

\setlength{\dbltextfloatsep}{12pt plus 2pt minus 2pt}
\begin{figure*}[!t]
    \centering
    \includegraphics[width=\textwidth,trim=0 6bp 0 5bp,clip]{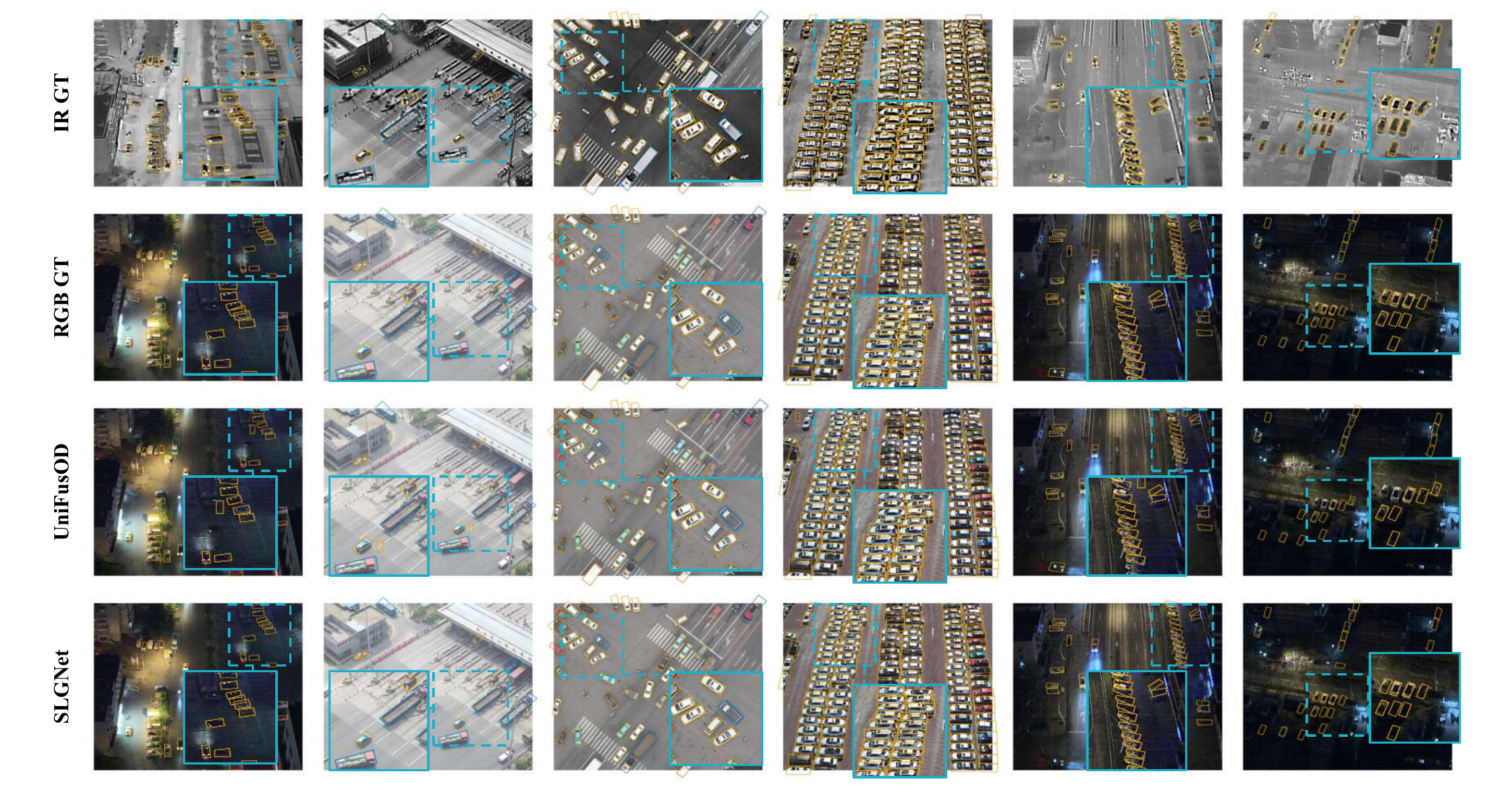}
    \caption{Detection results on DroneVehicle. The first two rows show ground-truth annotations on infrared and RGB images. The third and fourth rows show UniFusOD~\cite{xiang2025infrared} and SLGNet, respectively. Cyan boxes enlarge faint, densely packed vehicles.}
    \label{fig:vis_drone}
\end{figure*}

\Needspace*{2\baselineskip}
\noindent\textbf{Feature Responses.} Fig.~\ref{fig:lgm_vis} compares feature responses with and without LGM under low illumination.
With LGM, responses cover pedestrians and riders more coherently, including targets barely visible in RGB.
The unmodulated maps are more fragmented in the same regions.
The more complete target responses are consistent with scene-conditioned recalibration of the structure-enriched features.

\subsection{Parameter Efficiency}

We compare SLGNet's frozen-backbone adaptation with a full fine-tuning baseline on FLIR and DroneVehicle, as summarized in Table~\ref{tab:efficiency}.
Full fine-tuning updates the backbone, whereas SLGNet keeps its weights fixed.
Trainable parameters count the weights updated during training.

SLGNet reduces the trainable parameter count from 96.0M to 12.1M, approximately 87\% fewer than full fine-tuning.
At the same time, mAP increases from 43.6 to 45.1 on FLIR and from 53.5 to 57.2 on DroneVehicle.
The frozen-backbone configuration thus improves detection accuracy with substantially fewer trainable weights.

\noindent\textbf{Training Dynamics.} Fig.~\ref{fig:training_curve} compares the FLIR validation mAP trajectories of adapter tuning and full fine-tuning, complementing the final accuracy comparison in Table~\ref{tab:efficiency}.
Both configurations improve during the early epochs.
Adapter tuning reaches a high validation mAP within the first 10 epochs, and its mean curve remains above full fine-tuning during subsequent training.
Its later trajectory remains relatively stable.
Full fine-tuning continues to improve more gradually but finishes at a lower validation mAP.
The narrower shaded band for adapter tuning over most epochs indicates less variation in the reported validation scores.
Together, the curves and parameter comparison show that the frozen-backbone configuration uses fewer trainable weights while sustaining a higher validation mAP across the later training epochs, rather than obtaining an advantage only at the final epoch.

\setlength{\textfloatsep}{8pt plus 2pt minus 2pt}
\begin{table}[!t]
    \centering
    \small
    \setlength{\tabcolsep}{3pt}
    \caption{Trainable parameters and detection accuracy with full fine-tuning and SLGNet. $\Delta$ gives the relative parameter reduction and absolute AP gains.}
    \label{tab:efficiency}
    \begin{tabular}{l r rrrr}
    \toprule
    \multirow{2}{*}{\textbf{Tuning}} &
    \multirow{2}{*}{\shortstack{\textbf{Trainable}\\\textbf{Params}}} &
    \multicolumn{2}{c}{\textbf{FLIR}} &
    \multicolumn{2}{c}{\textbf{DroneVehicle}} \\
    \cmidrule(lr){3-4} \cmidrule(lr){5-6}
    & & $\mathrm{mAP}$ & $\mathrm{mAP}_{50}$ & $\mathrm{mAP}$ & $\mathrm{mAP}_{50}$ \\
    \midrule
    Full tuning & 96.0M & 43.6 & 82.2 & 53.5 & 75.3 \\
    \rowcolor[gray]{0.94} Adapter tuning & \textbf{12.1M} & \textbf{45.1} & \textbf{85.8} & \textbf{57.2} & \textbf{80.7} \\
    \midrule
    $\Delta$ & -87\% & +1.5 & +3.6 & +3.7 & +5.4 \\
    \bottomrule
    \end{tabular}
\end{table}

\enlargethispage{-\baselineskip}
\subsection{Qualitative Analysis}
Fig.~\ref{fig:vis_drone} compares detections on DroneVehicle, where low illumination, small object scales, dense layouts, and complex background textures pose challenges.
In the magnified regions, UniFusOD~\cite{xiang2025infrared} (third row) fails to distinguish adjacent vehicles or misses faint targets because weak structural signals are lost during fusion.
For these low-contrast and densely packed vehicle targets, SLGNet (fourth row) \mbox{produces} oriented bounding boxes that more closely match the corresponding ground-truth annotations.

This advantage can be attributed to the recovery of subtle high-frequency edge cues, such as vehicle contours, by the SA-Adapter, which helps separate densely packed vehicles.
Meanwhile, LGM suppresses background clutter to amplify weak target responses.
These visual results support our design rationale of combining structure-guided fusion with scene-conditioned recalibration and corroborate the quantitative oriented detection gains in Table~\ref{tab:DroneVehicle_comparison}.

\section{Conclusion}
We have presented SLGNet, a parameter-efficient RGB--infrared detection framework that combines structure-guided fusion with scene-conditioned feature recalibration.
The Structure-Aware Adapter uses a complementary edge reference to guide multi-scale fusion and injects the fused features into a frozen ViT.
LGM then uses structured scene descriptions to rescale and shift the final backbone features.
Experiments on four benchmarks show competitive performance, and component ablations show that scene conditioning adds gains beyond structural guidance.
Compared with our full fine-tuning baseline, SLGNet achieves the highest detection accuracy among all compared methods on FLIR and DroneVehicle with approximately 87\% fewer trainable parameters.
Future work will focus on reducing the computational cost of generating structured scene descriptions.

\FloatBarrier
\section*{LLM Usage Statement}
In preparing this manuscript, we used Gemini to refine the language and improve readability.
We used Codex to improve the layout and presentation and to assist in checking the reported data and references.

{\small
\bibliographystyle{ieeenat_fullname}
\bibliography{main}
}

\end{document}